\documentclass{article} 

\usepackage[accepted]{map/icml2022}

\usepackage{times}
\usepackage{amssymb}
\usepackage{graphicx}
\usepackage{wrapfig}
\usepackage{rotating}
\usepackage{caption}
\usepackage{mathtools}
\usepackage{listings}
\usepackage{natbib}
\usepackage{subcaption}
\usepackage{tablefootnote}
\usepackage{booktabs}
\usepackage{url}
\usepackage[automake, acronym, toc]{glossaries}
\usepackage[draft]{hyperref, bookmark}
\usepackage[capitalize,noabbrev]{cleveref}

\newglossary{symbols}{sym}{sbl}{List of Abbreviations and Symbols}
\makeglossaries
\newacronym{gd}{GD}{Gradient Descent}
\newacronym{sgd}{SGD}{Stochastic Gradient Descent}
\newacronym{sgdm}{SGDM}{Stochastic Gradient Descent with Momentum}
\newacronym{nsgdm}{N-SGDM}{Normalised \gls{sgdm}}
\newacronym{cnsgdm}{CN-SGDM}{Component Normalised \gls{sgdm}}
\newacronym{osgdm}{Orthogonal-SGDM}{Orthogonal Stochastic Gradient Descent with Momentum}
\newacronym{lars}{LARS}{Layer-wise Adaptive Rate Scaling}
\newacronym{svd}{SVD}{Singular Value Decomposition}
\newacronym{relu}{ReLU}{Rectified Linear Unit}
\newacronym{kkt}{KKT}{Karush-Kuhn-Tucker}
\newacronym{mlp}{MLP}{Multi-layered Perceptron}

\newglossaryentry{adam}{
    name=Adam,
    description={An optimiser that uses ADAptive Moment estimation}
}

\newglossaryentry{oadam}{
    name=Orthogonal-Adam,
    description={An orthogonalised version of ADAptive Moment estimation}
}

\DeclarePairedDelimiter\abs{\lvert}{\rvert}
\DeclarePairedDelimiter\norm{\lVert}{\rVert}
\makeatletter
\let\oldabs\abs
\def\abs{\@ifstar{\oldabs}{\oldabs*}}
\let\oldnorm\norm
\def\norm{\@ifstar{\oldnorm}{\oldnorm*}}
\makeatother

\newcommand{\av}[2][]{\mathbb{E}_{#1\!}\left[ #2 \right]}

\newcommand{\tr}{\mathsf{T}}
\renewcommand{\vec}[1]{#1}
\newcommand{\mat}[2][]{%
    \expandafter\MakeUppercase\expandafter{#2}
    \ifthenelse{\equal{#1}{inv}}{^{-1}}{}
    \ifthenelse{\equal{#1}{t}}{^\tr}{}
}

\newcommand{\reals}[1]{\mathop{\mathbb{R}}\ifthenelse{\equal{#1}{}}{}{^{#1}}}
\newcommand{\nonnegint}{\mathop{\mathbb{Z}^{0+}}}

\newcommand{\w}{\theta}

\newcommand{\ie}{\textit{i.e.}\ }
\newcommand{\eg}{\textit{e.g.}\ }

\icmltitlerunning{Gradient Orthogonalisation}

\begin{document}

\twocolumn[
	\icmltitle{Orthogonalising gradients to speedup neural network optimisation}



	\icmlsetsymbol{equal}{*}

	\begin{icmlauthorlist}
		\icmlauthor{Mark Tuddenham}{soton}
		\icmlauthor{Adam Prügel-Bennett}{soton}
		\icmlauthor{Jonathon Hare}{soton}
	\end{icmlauthorlist}

	\icmlaffiliation{soton}{Department of Electronics and Computer Science, University of Southampton, Southampton, UK}

	\icmlcorrespondingauthor{Mark	Tuddenham}{mark.tuddenham@southampton.ac.uk}

	\icmlkeywords{Machine Learning, ICML}

	\vskip 0.3in
]



\printAffiliationsAndNotice{}  



\begin{abstract}
	The optimisation of neural networks can be sped up by orthogonalising the gradients before the optimisation step, ensuring the diversification of the learned representations.
	We orthogonalise the gradients of the layer's components/filters with respect to each other to separate out the intermediate representations.
	Our method of orthogonalisation allows the weights to be used more flexibly, in contrast to restricting the weights to an orthogonalised sub-space.
	We tested this method on ImageNet and CIFAR-10 resulting in a large decrease in learning time, and also obtain a speed-up on the semi-supervised learning BarlowTwins.
	We obtain similar accuracy to SGD without fine-tuning and better accuracy for naïvely chosen hyper-parameters.
\end{abstract}

\section{Introduction}\label{sec:intro}

\begin{figure}[!ht]
	\centering
	\includegraphics[width=0.9\linewidth]{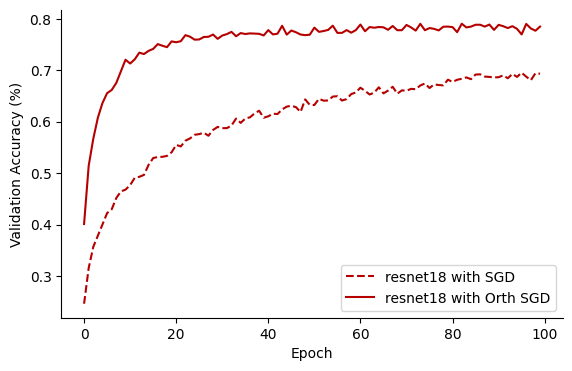}
	\caption{An example of the speed-up obtained by orthogonalising the gradients on CIFAR-10.}\label{fig:example_speed_up}
\end{figure}

Neural network layers are made up of several identical, but differently parametrised, components, \eg filters in a convolutional layer, or heads in a multi-headed attention layer.
Layers consist of several components so that they can provide a diverse set of intermediary representations to the next layer, however, there is no constraint or bias, other than the implicit bias from the cost function, to learning different parametrisations.
We introduce this diversification bias in the form of orthogonalised gradients and find a resultant speed-up in learning and sometimes improved performance, see \cref{fig:example_speed_up}.

Our novel contributions include this new optimisation method, thorough testing on CIFAR-10 and ImageNet, additional testing on a semi-supervised learning method, and experiments to support our hypothesis.

In \cref{sec:overview} we detail the method and results to give an understanding of how this method works and its capabilities.
Then, in \cref{sec:discussion}, we provide experimental justifications and supporting experiments for this method along with finer details of the implementation and limitations.

\section{Overview of new method and results}\label{sec:overview}

\subsection{Related works}

Gradient orthogonalisation has been explored in the domain of multi-task learning~\citep{yu2020gradient} to keep the different tasks separate and relevant.
However in this work we focus on orthogonalisation for improving single task performance.

Weight orthogonalisation has been extensively explored with both empirical~\citep{bansal2018can,jia2017improving} and theoretical~\citep{jia2019orthogonal} justifications.
However, modifying the weights during training is unstable, and, in addition, it limits the weights to a tiny subspace.
Deep learning is know to work well despite the immense size of the weight space, and as such we do not view this as an advantage.
\citet{xie2017all} obtain improved performance over \gls{sgd} via weight orthogonalisation and allows them to train very deep networks, we aim to achieve the same results while being more flexible with model and optimisation method choice.
We do this by orthogonalising the gradients before they are used by an optimisation method rather than modifying the weights themselves.
This, in effect, biases the training towards learning orthogonal representations and we show this to be the case; in many areas of deep learning, introducing a bias instead of a hard constraint is preferred.

\subsection{Orthogonalising Gradients}
Given a neural network, $f$, with $L$ layers made from components, $c$,
\begin{align}
	f      & = \circ^L_{i=1} \left(f_i\right),                                             \\
	f_l(x) & = [{c_l}_1(x), {c_l}_2(x), \dots, {c_l}_{N_l}(x)],\label{eq:components_model}
\end{align}
where $\circ$ is the composition operator, $N_l$ is the number of components in layer $l$, $c_l : \reals{S_{l-1}\times N_{l-1}}\rightarrow\reals{S_l}$ is a parametrised function and ${c_l}_i$ denotes $c_l$ parametrised with ${\w_l}_i\in\reals{P_l}$ giving $f_l : \reals{S_{l-1}\times N_{l-1}}\rightarrow\reals{S_l\times N_l}$ parametrised by $\w_l\in\reals{P_l\times N_l}$.

Let
\begin{equation}
	\mat{G}_l = [\nabla {c_l}_1, \nabla {c_l}_2, \dots, \nabla {c_l}_{N_l}],\label{eq:G_l}
\end{equation}
be the $P_l \times N_l$ matrix of the components' gradients.

Then the nearest orthonormal matrix, \ie the orthonormal matrix, $\mat{O}_l$, that minimises the Frobenius norm of its difference from $\mat{G}_l$
\[
	\min_{\mat{O}_l} \norm{\mat{O}_l - \mat{G}_l} \quad \text{subject to} \:\; \forall i,j : \left\langle {\mat{O}_l}_i, {\mat{O}_l}_j \right\rangle = \delta_{ij},
\]
where $\delta_{ij}$ is the Kronecker delta function, is the product of the left and right singular vector matricies from the \gls{svd} of $\mat{G}_l$~\citep{trefethen1997numerical},
\begin{align}
	\mat{G}_l & = \mat{U}_l \mat{\Sigma}_l \mat{V}_l^\tr, \\
	\mat{O}_l & = \mat{U}_l \mat{V}_l^\tr.
\end{align}
Thus, we can adjust a first-order gradient descent method, such as \gls{sgdm}~\citep{polyak1964some}, to make steps where the components are pushed in orthogonal directions,
\begin{align}
	v_l^{(t+1)}  & = \gamma v_l^{(t)} + \eta \mat{O}_l^{(t)}, \; \text{and} \\
	\w_l^{(t+1)} & = \w_l^{(t)} - v_l^{(t+1)},
\end{align}
where $v_l$ is the velocity matrix, $t\in\nonnegint$ is the time, $\gamma$ is the momentum decay term, and $\eta$ is the step size.
We call this method \gls{osgdm}.
This modification can clearly be applied to any first-order optimisation algorithm by replacing the gradients with $\mat{O}^{(t)}_l$ before the calculation of the next iterate.

	Code for creating orthogonal optimisers in PyTorch is provided at \url{https://github.com/MarkTuddenham/Orthogonal-Optimisers}.
	And code for the experiments in this work is provided at \url{https://github.com/MarkTuddenham/Orthogonalised-Gradients}

\subsection{Results}

\subsubsection{CIFAR-10}\label{sec:cifar10}

\begin{figure}[!ht]
	\centering
	\includegraphics[width=0.9\linewidth]{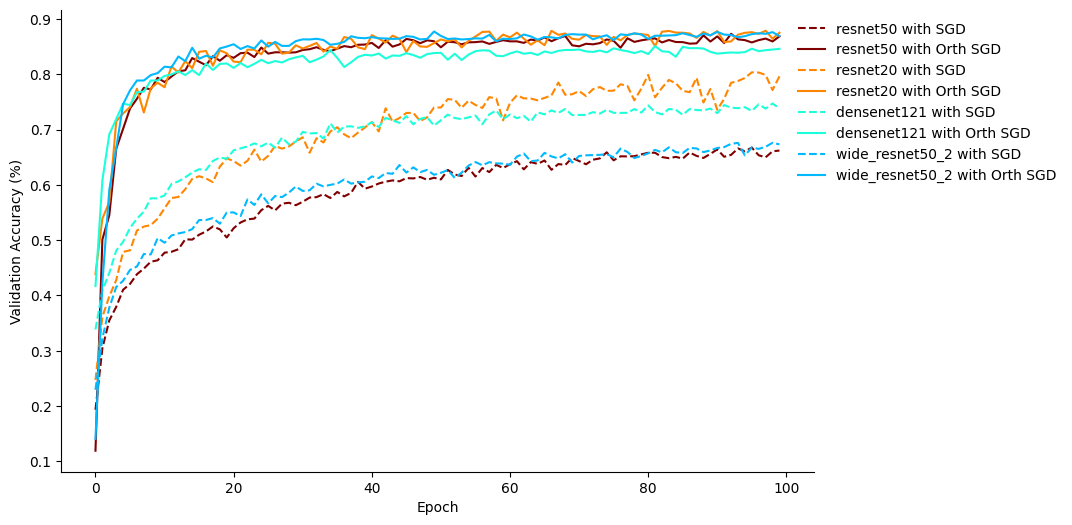}
	\caption{Validation accuracy from one run of \acrshort{sgdm} vs \acrshort{osgdm} for a selection of models. Full plot in \cref{sec:full_orth}. Best viewed in colour.}\label{fig:sgd_vs_orth_sgd_acc}
\end{figure}

\begin{figure}[!ht]
	\centering
	\includegraphics[width=0.9\linewidth]{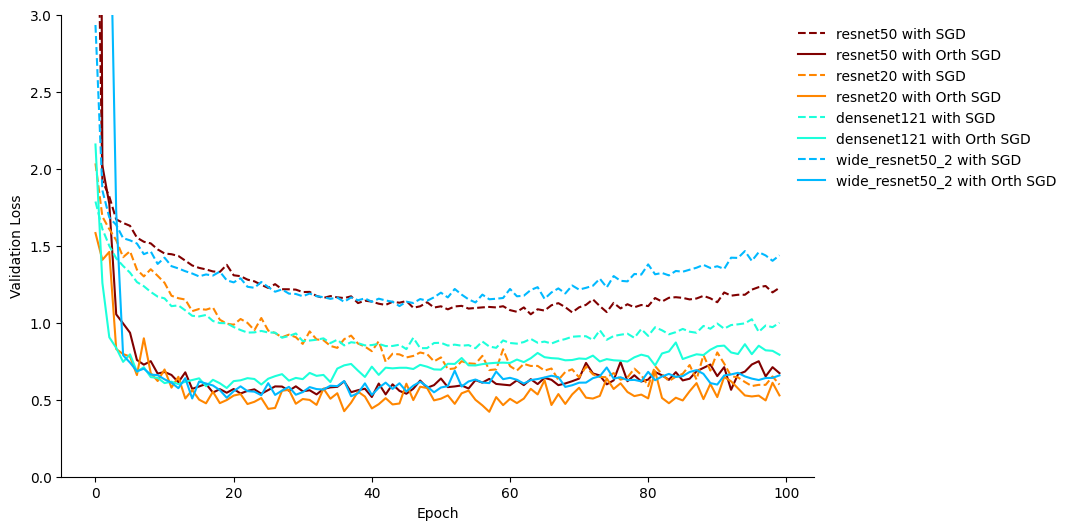}
	\caption{Validation losses from one run of \acrshort{sgdm} vs \acrshort{osgdm} for a selection of models. Full plot in \cref{sec:full_orth}. Best viewed in colour.}\label{fig:sgd_vs_orth_sgd_loss}
\end{figure}

We trained a suite of models on the CIFAR-10~\citep{krizhevsky2009learning} data set with a mini-batch size of 1024, learning rate of $10^{-2}$, momentum of 0.9, and a weight decay of $5\times 10^{-4}$ for 100 epochs.
We then repeated this using \gls{osgdm} instead of \gls{sgdm} and plot the results in \cref{fig:sgd_vs_orth_sgd_acc,fig:sgd_vs_orth_sgd_loss} and \cref{tab:orth_cifar_results}.

\begin{table}[ht]
	\centering
	\tiny
	\caption{Test loss and accuracy across a suite of models on CIFAR-10 comparing normal \acrshort{sgdm} with \acrshort{osgdm}, standard error across five runs.}\label{tab:orth_cifar_results}
	\begin{tabular}{lcccc}
		                                                                              & \multicolumn{2}{c}{Test Loss} & \multicolumn{2}{c}{Test Accuracy (\%)}                                                \\
		\cmidrule(lr){2-3}\cmidrule(lr){4-5}
		                                                                              & \acrshort{sgdm}               & \acrshort{osgdm}                       & \acrshort{sgdm}  & \acrshort{osgdm}          \\ \midrule
		BasicCNN\tablefootnote{As described in \cref{sec:BasicCNN}}                   & 0.7603 $\pm$ 0.0061           & 0.6808 $\pm$ 0.0038                    & 73.60 $\pm$ 0.19 & \textbf{76.67} $\pm$ 0.10 \\
		resnet20\tablefootnote{Model same as in \citet{he2015deep}}                   & 0.6728 $\pm$ 0.0301           & 0.6766 $\pm$ 0.0155                    & 79.14 $\pm$ 0.62 & \textbf{87.12} $\pm$ 0.12 \\
		resnet44\footnotemark[\value{footnote}]                                       & 0.7000 $\pm$ 0.0166           & 0.7600 $\pm$ 0.0299                    & 79.81 $\pm$ 0.37 & \textbf{88.12} $\pm$ 0.20 \\
		resnet18\tablefootnote{From \url{https://pytorch.org/vision/0.9/models.html}} & 0.9656 $\pm$ 0.0104           & 0.8427 $\pm$ 0.0121                    & 77.01 $\pm$ 0.21 & \textbf{84.68} $\pm$ 0.12 \\
		resnet34\footnotemark[\value{footnote}]                                       & 1.0468 $\pm$ 0.0134           & 0.7087 $\pm$ 0.0165                    & 75.86 $\pm$ 0.26 & \textbf{85.42} $\pm$ 0.33 \\
		resnet50\footnotemark[\value{footnote}]                                       & 1.2304 $\pm$ 0.0462           & 0.6797 $\pm$ 0.0235                    & 67.99 $\pm$ 0.73 & \textbf{86.51} $\pm$ 0.12 \\
		densenet121\footnotemark[\value{footnote}]                                    & 1.0027 $\pm$ 0.0132           & 0.8669 $\pm$ 0.0132                    & 75.26 $\pm$ 0.30 & \textbf{84.34} $\pm$ 0.15 \\
		densenet161\footnotemark[\value{footnote}]                                    & 1.1399 $\pm$ 0.0096           & 1.1688 $\pm$ 0.1960                    & 75.81 $\pm$ 0.20 & \textbf{85.51} $\pm$ 0.19 \\
		resnext50\_32x4d\footnotemark[\value{footnote}]                               & 1.2470 $\pm$ 0.0254           & 0.6669 $\pm$ 0.0223                    & 68.73 $\pm$ 0.30 & \textbf{86.37} $\pm$ 0.24 \\
		wide\_resnet50\_2\footnotemark[\value{footnote}]                              & 1.4141 $\pm$ 0.0337           & 0.7018 $\pm$ 0.0091                    & 69.42 $\pm$ 0.33 & \textbf{87.30} $\pm$ 0.12 \\
	\end{tabular}
\end{table}

\begin{figure}[!ht]
	\centering
	\includegraphics[width=0.9\linewidth]{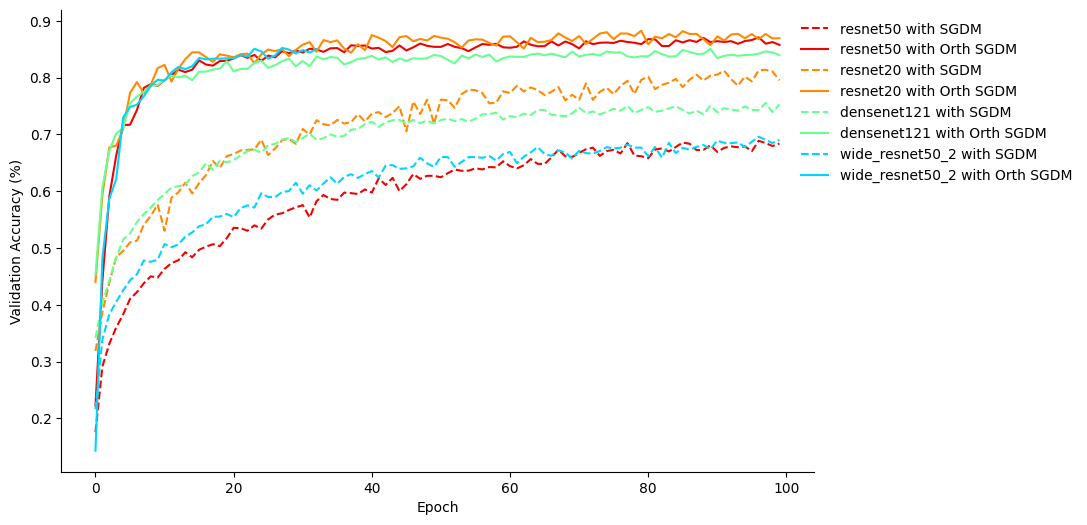}
	\caption{Train accuracy from one run of \acrshort{sgdm} vs \acrshort{osgdm} for a selection of models. Best viewed in colour.}\label{fig:sgd_vs_orth_sgd_acc_train}
\end{figure}

\begin{figure}[!ht]
	\centering
	\includegraphics[width=0.9\linewidth]{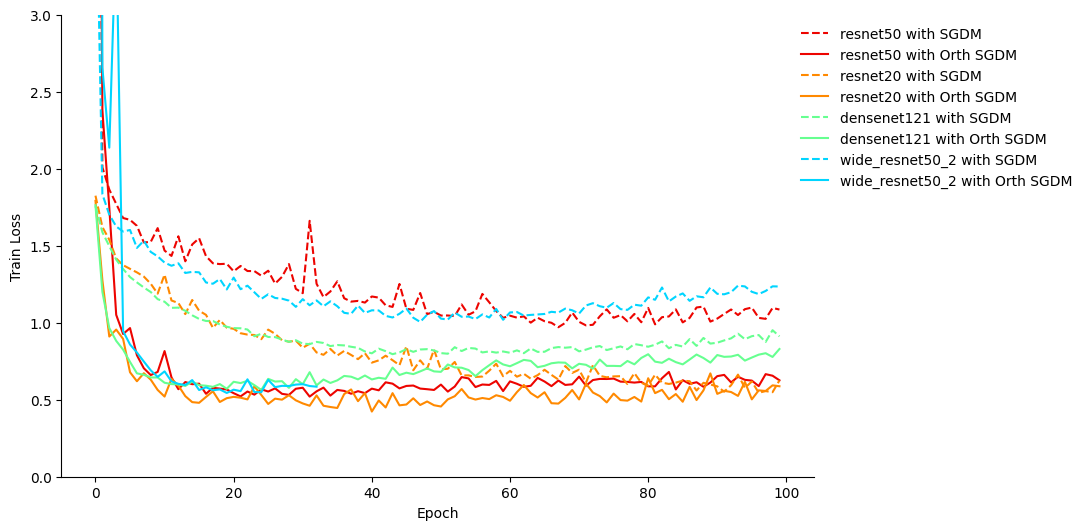}
	\caption{Train losses from one run of \acrshort{sgdm} vs \acrshort{osgdm} for a selection of models. Best viewed in colour.}\label{fig:sgd_vs_orth_sgd_loss_train}
\end{figure}

\Gls{osgdm} is more efficient and achieves better test accuracy than \gls{sgdm} for every model we trained on CIFAR-10 without hyper-parameter tuning.
The validation curves follow the training curves, \cref{fig:sgd_vs_orth_sgd_acc_train,fig:sgd_vs_orth_sgd_loss_train}, and have the same patterns, this means that \gls{osgdm} exhibits the same generalisation performance as \gls{sgdm}.
More importantly though, we can see that the model learns much faster at the beginning of training, as shown by \cref{fig:sgd_vs_orth_sgd_acc}, this means that we do not need as many epochs to get to a well-performing network.
This is especially good in light of the large data sets that new models are being trained on, where they are trained for only a few epochs, or even less~\citep{brown2020language}.

For \gls{sgdm} the performance of the residual networks designed for ImageNet~\citep{deng2009imagenet} (18, 34, 50) get worse as the models get bigger.
The original ResNet authors, \citet{he2015deep}, note that unnecessarily large networks may over-fit on a small data set such as CIFAR-10.
However, when trained with \gls{osgdm}, these models do not suffer from this over-parametrisation problem and even slightly improve in performance as the models get bigger, in clear contrast to \gls{sgdm}.
This agnosticism to over-parametrisation helps alleviate the need for the practitioner to tune a model's architecture to the task at hand to achieve a reasonable performance.

\subsubsection{In comparison to Adam}
We compare our method to the \gls{adam} optimiser~\citep{kingma2014adam}.
\Gls{adam} has found its place as a reliable optimiser that works over a wide variety of hyper-parameter sets, yielding consistent performance with little fine-tuning needed.
We see our optimisation method occupying the same space as \gls{adam}.

\begin{table}[th]
	\tiny
	\centering
	\caption{Test accuracy across a suite of hyper-parameter sets on CIFAR-10 on a resnet20, standard error across five runs. For \gls{adam} $\beta_2 = 0.99$.}\label{tab:orth_cifar_results_adam_acc}
	\begin{tabular}{llcccc}
		          &          & \multicolumn{2}{c}{\gls{sgdm}} & \multicolumn{2}{c}{\gls{adam}}                                                                      \\
		\cmidrule(lr){3-4}\cmidrule(lr){5-6}
		LR        & Mom & Original                       & Orthogonal                         & Original                  & Orthogonal                         \\ \midrule

		$10^{-1}$ & 0.95     & 83.59 {\tiny $\pm$ 2.09 }      & \textbf{85.58} {\tiny $\pm$ 0.98 } & 38.95 {\tiny $\pm$ 8.95 } & 76.84 {\tiny $\pm$ 1.53 }          \\
		$10^{-2}$ & 0.95     & 82.66 {\tiny $\pm$ 1.02 }      & \textbf{87.72} {\tiny $\pm$ 0.44 } & 74.23 {\tiny $\pm$ 2.38 } & 86.48 {\tiny $\pm$ 0.17 }          \\
		$10^{-3}$ & 0.95     & 66.59 {\tiny $\pm$ 0.44 }      & \textbf{85.88} {\tiny $\pm$ 0.33 } & 83.08 {\tiny $\pm$ 0.76 } & 85.12 {\tiny $\pm$ 0.06 }          \\
		\midrule
		$10^{-1}$ & 0.9      & 82.52{\tiny $\pm$ 1.16}        & \textbf{85.06}{\tiny $\pm$ 0.47}   & 28.26{\tiny $\pm$ 7.16}   & 73.62{\tiny $\pm$ 2.96}            \\
		$10^{-2}$ & 0.9      & 79.96{\tiny $\pm$ 0.48}        & \textbf{87.44}{\tiny $\pm$ 0.25}   & 73.46{\tiny $\pm$ 1.19}   & 85.26{\tiny $\pm$ 0.38}            \\
		$10^{-3}$ & 0.9      & 60.69{\tiny $\pm$ 0.18}        & 84.67{\tiny $\pm$ 0.21}            & 83.16{\tiny $\pm$ 0.66}   & \textbf{85.25}{\tiny $\pm$ 0.31}   \\
		\midrule
		$10^{-1}$ & 0.8      & 84.16 {\tiny $\pm$ 0.43 }      & \textbf{86.01} {\tiny $\pm$ 0.74 } & 27.50 {\tiny $\pm$ 6.76 } & 71.88 {\tiny $\pm$ 4.25 }          \\
		$10^{-2}$ & 0.8      & 77.42 {\tiny $\pm$ 0.98 }      & \textbf{87.18} {\tiny $\pm$ 0.12 } & 72.60 {\tiny $\pm$ 1.76 } & 86.75 {\tiny $\pm$ 0.26 }          \\
		$10^{-3}$ & 0.8      & 53.21 {\tiny $\pm$ 0.43 }      & 82.95 {\tiny $\pm$ 0.40 }          & 80.89 {\tiny $\pm$ 1.93 } & \textbf{85.52} {\tiny $\pm$ 0.29 } \\
		\midrule
		$10^{-1}$ & 0.5      & 80.08 {\tiny $\pm$ 0.36 }      & \textbf{87.37} {\tiny $\pm$ 0.18 } & 18.39 {\tiny $\pm$ 4.84 } & 72.10 {\tiny $\pm$ 2.83 }          \\
		$10^{-2}$ & 0.5      & 68.64 {\tiny $\pm$ 1.05 }      & \textbf{86.05} {\tiny $\pm$ 0.10 } & 71.62 {\tiny $\pm$ 1.95 } & 84.22 {\tiny $\pm$ 0.69 }          \\
		$10^{-3}$ & 0.5      & 43.51 {\tiny $\pm$ 1.02 }      & 78.68 {\tiny $\pm$ 0.77 }          & 81.67 {\tiny $\pm$ 1.05 } & \textbf{84.21} {\tiny $\pm$ 0.43 } \\
	\end{tabular}
\end{table}

Our method outperforms \gls{adam} on all but one hyper-parameter set --- \cref{tab:orth_cifar_results_adam_acc}.
In addition since we can apply our method to any previous optimisation method, we also test \gls{oadam} and find that it outperforms \gls{adam} too including at high learning rates where \gls{adam} suffers from blow-ups.
See \cref{fig:adam1,fig:adam2} for the training plots.

\subsubsection{Matching ResNet's performance}\label{sec:match_resnet}

Having shown that \gls{osgdm} speeds up learning with non-optimised hyper-parameters, we now aim to show that it can achieve state-of-the-art results.
To do this we use the same hyper-parameters as the original ResNet paper~\citep{he2015deep}, which have been painstakingly tuned to benefit \gls{sgdm}, to train using \gls{osgdm}.

\begin{table}[t]
	\small
	\centering
	\caption{Test loss and accuracy of a resnet20, as in \citet{he2015deep}, on CIFAR-10; hyper-parameter tuned to normal \acrshort{sgdm} vs \acrshort{osgdm}, standard error across five runs. Mini-batch size of 128, see \cref{sec:limitations} for why this hyper-parameter value impedes \gls{osgdm}, learning-rate of 0.1, momentum of 0.9, weight-decay of $10^{-4}$, and a learning rate schedule of $\times 0.1$ at epochs 100, 150 for 200 epochs.} \label{tab:sgd_vs_orth_sgd_resnet_tuned}
	\begin{tabular}{lcc}
		                                   & Test Loss           & Test Acc (\%) \\ \midrule
		\acrshort{sgdm}~\citep{he2015deep} & ---                 & \textbf{91.25}     \\
		\acrshort{sgdm}                    & 0.4053 $\pm$ 0.0054 & 91.17 $\pm$ 0.28   \\
		\acrshort{osgdm}                   & 0.4231 $\pm$ 0.0043 & 90.18 $\pm$ 0.30   \\
	\end{tabular}
\end{table}

This also tests the efficacy of \gls{osgdm} as a drop-in replacement for \gls{sgdm}.
\Gls{osgdm} gets close to the original results, \cref{tab:sgd_vs_orth_sgd_resnet_tuned}, even though the hyper-parameters are perfected for \gls{sgdm}.
It is the authors' belief that with enough hyper-parameter tuning \gls{sgd} or \gls{sgdm} will be the best optimisation method; however, this experiment shows that \gls{osgdm} is robust to hyper-parameter choice and can easily replace \gls{sgdm} in existing projects.
Unfortunately, the authors do not have the compute-power to extensively hyper-parameter tune a residual network for \gls{osgdm}, however, it is exceedingly likely that better results would be achieved by doing so.

\subsubsection{ImageNet}
\begin{figure}[!ht]
	\centering
	\includegraphics[width=0.9\linewidth]{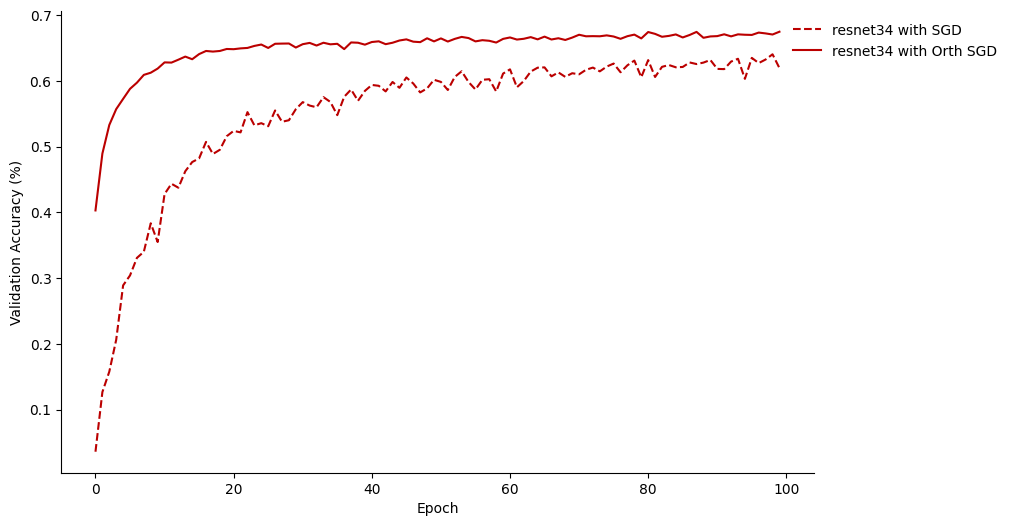}
	\caption{Validation accuracy of \acrshort{sgdm} vs \acrshort{osgdm} on ImageNet}\label{fig:sgd_vs_orth_sgd_imgnet}
\end{figure}
\Gls{osgdm} also works on a large data set such as ImageNet~\citep{deng2009imagenet} --- \cref{fig:sgd_vs_orth_sgd_imgnet}.
Using a resnet34, mini-batch size of 1024, learning rate of $10^{-2}$, momentum of 0.9, and a weight decay of $5\times 10^{-4}$, for 100 epochs.
\gls{sgdm} achieves a test accuracy of 61.9\% and a test loss of 1.565 while \gls{osgdm} achieves 67.5\% and 1.383 respectively.
While these results are a way off the capabilities of the model they still demonstrate a significant speed-up and improvement from using \gls{osgdm}, especially at the start of learning, and further reinforces how a dearth of hyper-parameter tuning impedes performance.

\subsubsection{Barlow Twins}
\begin{figure}[!ht]
	\centering
	\includegraphics[width=0.9\linewidth]{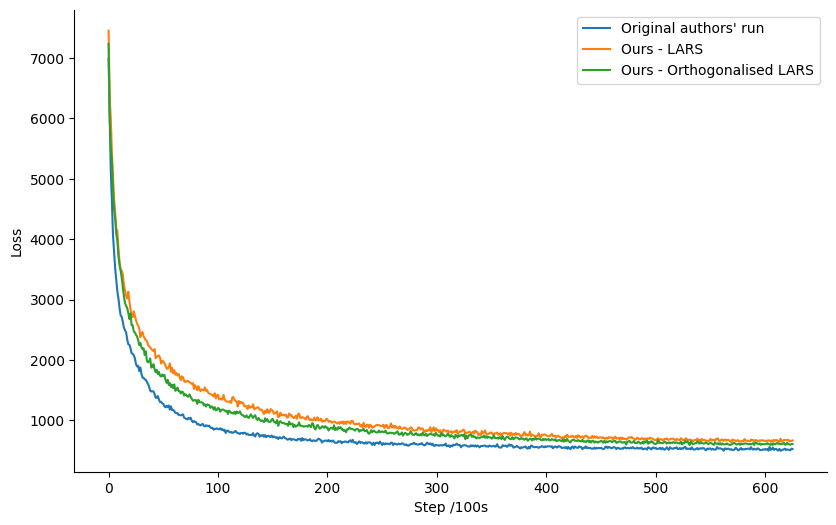}
	\caption{Barlow Twins loss during the unsupervised phase using \acrshort{lars} and Orthogonal \acrshort{lars} on ImageNet}\label{fig:barlow_twins_losses}
\end{figure}

Barlow Twins~\citep{zbontar2021barlow} is a semi-supervised method that uses ``the cross-correlation matrix between the outputs of two identical networks fed with distorted versions of a sample'' to avoid collapsing to trivial solutions.
While the authors do provide code, we could not replicate their results by running it.
To train within our compute limitations we used a mini-batch size of 1024 instead of 2048 however this should not affect the results since ``Barlow Twins does not require large batches''~\citep{zbontar2021barlow}.
Additionally, Barlow Twins uses the \gls{lars} algorithm~\citep{you2017large}, which is designed to adjust the learning rate based on the ratio between the magnitudes of the gradients and weights, there should be no significant slow-down, or speed-up, in learning due to the mini-batch size.
We do not orthogonalise the gradients for the dense layers (see \cref{sec:orth_conv_vs_dense}).

Comparing our own runs, we establish that orthogonalising the gradients before the \gls{lars} algorithm does speed up learning as shown in \cref{fig:barlow_twins_losses}, in agreement with previous experiments.
This is evidence that orthogonalising gradients is also beneficial for semi-supervised learning and, moreover, that optimisation algorithms other than \gls{sgdm} can be improved in this way.

\section{Discussion of problem and method}\label{sec:discussion}

\subsection{Normalisation}
When we perform \gls{svd} on the reshaped gradient tensor, we obtain an orthonormal matrix, since this changes the magnitude of the gradient we look at the effect of this normalisation.
\Gls{nsgdm}~\citep{nesterov2003introductory} provides an improvement in non-convex optimisation since it is difficult to get stuck in a local minimum as the step size is not dependent on the gradient magnitude.
However, it hinders convergence to a global minimum since there is no way of shortening the step size; deep learning is highly non-convex and is unlikely to be optimised to a global minimum.
Therefore, it stands to reason that normalising the gradient would speed up the optimisation of deep networks.

We compare \gls{nsgdm} to normalising the gradients per component --- \ie normalising the columns of $G_l$, \cref{eq:G_l}, instead of orthognormalising it --- \gls{cnsgdm}, as well as to \gls{sgdm} and \gls{osgdm}.
\Gls{nsgdm} improves over \gls{sgdm}, and \gls{cnsgdm} improves over \gls{nsgdm} except from the oft case where it diverges.
Finally, \gls{osgdm} obtains the best solutions while remaining stable on all the models.

\begin{table}[ht]
	\centering
	\footnotesize
	\setlength\tabcolsep{2pt}
	\caption{Test accuracy for several models trained with \gls{sgdm}, \acrlong{nsgdm}, \acrlong{cnsgdm}, and \gls{osgdm}; trained as in \cref{sec:cifar10}.}\label{tab:normalised_sgdm_acc}
	\begin{tabular}{lcccc}
		            & \acrshort{sgdm}         & \acrshort{nsgdm}        & \acrshort{cnsgdm}       & \acrshort{osgdm}                 \\ \midrule
		BasicCNN    & 73.68{\tiny $\pm$ 0.27} & 73.72{\tiny $\pm$ 0.45} & 74.53{\tiny $\pm$ 0.32} & \textbf{76.75}{\tiny $\pm$ 0.23} \\
		resnet18    & 76.83{\tiny $\pm$ 0.22} & 78.94{\tiny $\pm$ 0.19} & 0.00{\tiny $\pm$ 0.00}  & \textbf{84.94}{\tiny $\pm$ 0.10} \\
		resnet50    & 69.35{\tiny $\pm$ 0.30} & 79.35{\tiny $\pm$ 0.21} & 0.00{\tiny $\pm$ 0.00}  & \textbf{86.59}{\tiny $\pm$ 0.10} \\
		resnet44    & 79.73{\tiny $\pm$ 1.27} & 83.60{\tiny $\pm$ 0.77} & 84.44{\tiny $\pm$ 0.55} & \textbf{87.49}{\tiny $\pm$ 0.39} \\
		densenet121 & 75.45{\tiny $\pm$ 0.20} & 79.06{\tiny $\pm$ 0.04} & 0.00{\tiny $\pm$ 0.00}  & \textbf{84.86}{\tiny $\pm$ 0.07} \\
	\end{tabular}
\end{table}

\subsection{Diversified intermediary representations}
Along with different parametrisations we also desire different intermediary representations, a model will perform better if its layers output $N$ different representations as opposed to $N$ similar ones.

Given $x_l$ are the resulting representations from the intermediary layers,
\[
	x_l = \left(\circ^l_{i=1} f_i\right) (x_0)
\]
where $x_0$ is the input and $x_l$ is the intermediary representation after layer $l$.
Then ${x_l}_i$ is the representation provided by ${c_l}_i$.

\begin{figure}[!ht]
	\centering
	\includegraphics[width=0.9\linewidth]{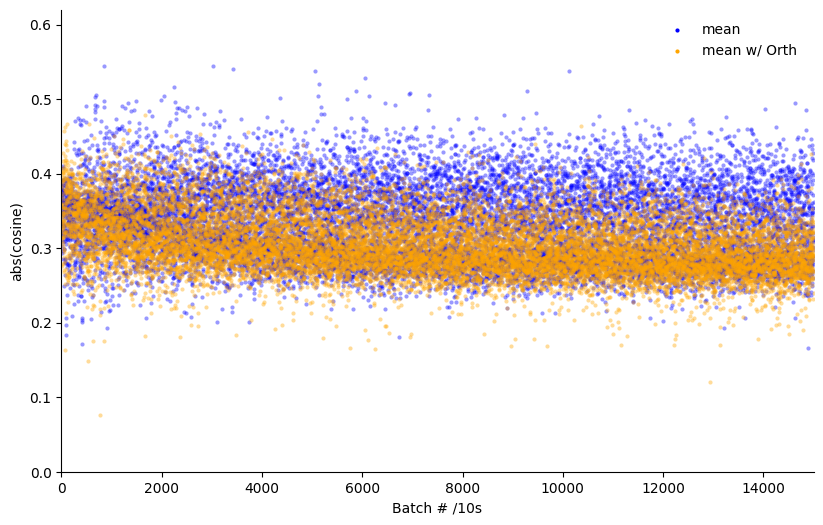}
	\caption{Mean of the absolute cosine of all distinct pairs of different intermediary representations, $\av{R_l}, \, l \in  \left\{1,2,3\right\}$, for all layers of a BasicCNN trained on CIFAR-10 as in \cref{sec:cifar10}.}\label{fig:R_l}
\end{figure}

We now look at the statistics of the absolute cosine of all distinct pairs of different latent features,
\[
	R_l = \left\{\abs{\langle {x_l}_i, {x_l}_j \rangle_2} \; | \; i<j \right\}.
\]
The representations have smaller cosines when using \gls{osgdm} versus \gls{sgdm} --- \cref{fig:R_l}.
In addition, \gls{osgdm} shows a steady decline in cosine similarity throughout training.
This indicates that more information is being passed to the next layer as the network is trained.

\subsection{Dead parameters}
\begin{figure}[!ht]
	\centering
	\begin{subfigure}{.9\linewidth}
		\centering
		\includegraphics[width=1\linewidth]{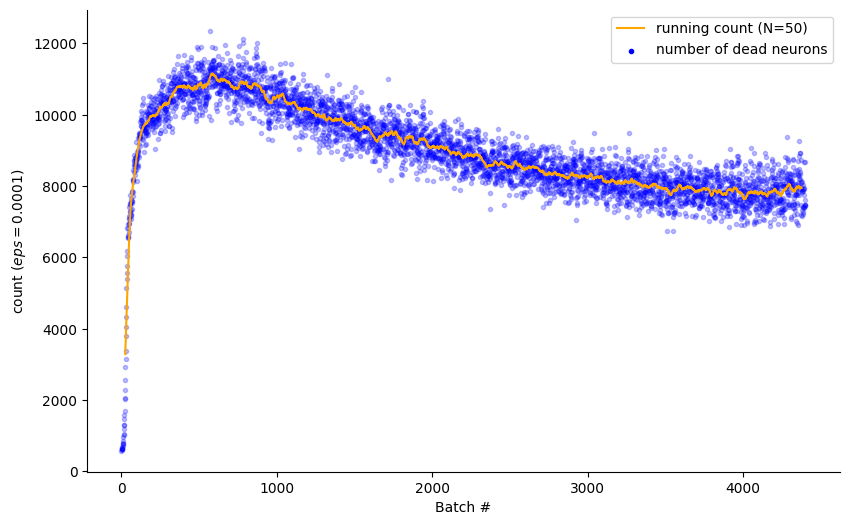}
		\caption{With \acrshort{sgdm}.}\label{fig:dead_params_sgdm}
	\end{subfigure}
	\begin{subfigure}{.9\linewidth}
		\centering
		\includegraphics[width=1\linewidth]{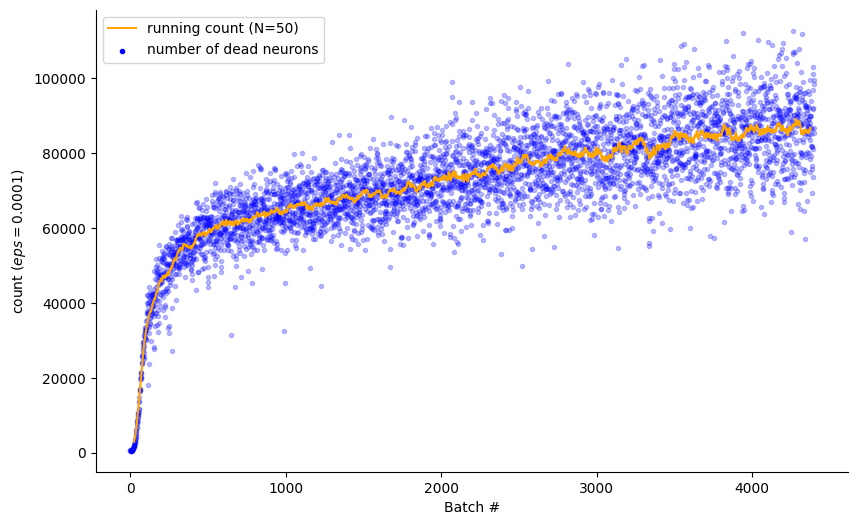}
		\caption{With \acrshort{osgdm}.}\label{fig:dead_params_osgdm}
	\end{subfigure}
	\caption{Number of temporarily dead parameters in layer2[1].conv2 of a resnet50 trained as in \cref{sec:cifar10}.}
	\label{fig:dead_params}
\end{figure}

Dead parameters occur when the activation function has a part with zero gradient, \eg a \gls{relu}.
If the result of the activation remains in this part, then the gradients of the preceding parameters will be zero and prevented from learning.
This limits the model's capacity based on a parameterisation, however temporarily dead parameters can be beneficial and act as a regulariser, similar to dropout.
To detect temporarily dead parameters, we simply look for parameters with zero gradient.
Comparing the amount of dead parameters produced by \gls{sgdm} versus \gls{osgdm}, \cref{fig:dead_params_sgdm,fig:dead_params_osgdm} respectively, shows that \gls{osgdm} ends with around and order of magnitude more temporarily dead parameters.
This implies a much higher regularisation which helps to explain \gls{osgdm}'s insensitivity to over-parametrisation.

\subsection{Implementation details}\label{sec:implementation}
While the QR decomposition is the most common orthogonalisation method, it is, in practice, less stable as the gradients are rank deficient~\citep[Section~3.5]{demmel1997applied}, \ie they have at least one small singular value.
\Gls{osgdm} has a longer wall time than \gls{sgdm} because of the added expense of the \gls{svd} which has non-linear time complexity in the matrix size.
In practice, we have found that the calculation of the \gls{svd} is either more than made up for by the speed up in iterates or a prohibitively expensive cost, with dense layers being the largest and so most problematic.

While there exist methods for computing an approximate \gls{svd} which are faster, we have used PyTorch's default implementation since we are more concerned with \gls{osgdm}'s performance and efficiency in iterates and not in wall time.
Even so the overhead is small, training a resnet20 as in \cref{sec:cifar10} takes 720.3 seconds with 96.4 of them taken up by the \gls{svd} calculation --- an increase of 15.5\% over normal \gls{sgdm}.
While this is a significant amount of time we can see that our method can take only 2\% of the number of epochs to reach the same accuracy --- \cref{fig:sgd_vs_orth_sgd_acc}.

It is doubtful that convergence of \gls{svd} is needed, so a custom matrix orthogonalisation algorithm, that has the required stability but remains fast and approximate, will reduce the computation overhead significantly and may allow previously infeasible networks to be optimised using \gls{osgdm}.
However, we note that even with a more suitable implementation, this method would still bias towards many smaller layers for a deeper, thinner network.

\subsection{Fully connected layers}\label{sec:orth_conv_vs_dense}

Fully connected or dense layers also fit our component model from \cref{eq:components_model} where the components are based on the inner product of the input and the parametrisation,
\[
	{c_l}_i(x) = \sigma(\langle \text{flatten}(x), {\w_l}_i \rangle),
\]
where $\sigma$ is an activation function, $S_l = 1$ giving $f_l : \reals{S_{l-1} \times N_{l-1}}\rightarrow \reals{N_l}$ and $\w_l \in \reals{S_{l-1}\cdot N_{l-1} \times N_l}$ as desired~\cite{wang2020orthogonal}.
Intuitively, each column of the weight matrix acts as a linear map resulting in one item in the output vector.
Thus, the gradients of fully connected layers can also be orthogonalised.

\begin{figure}[!ht]
	\centering
	\includegraphics[width=0.9\linewidth]{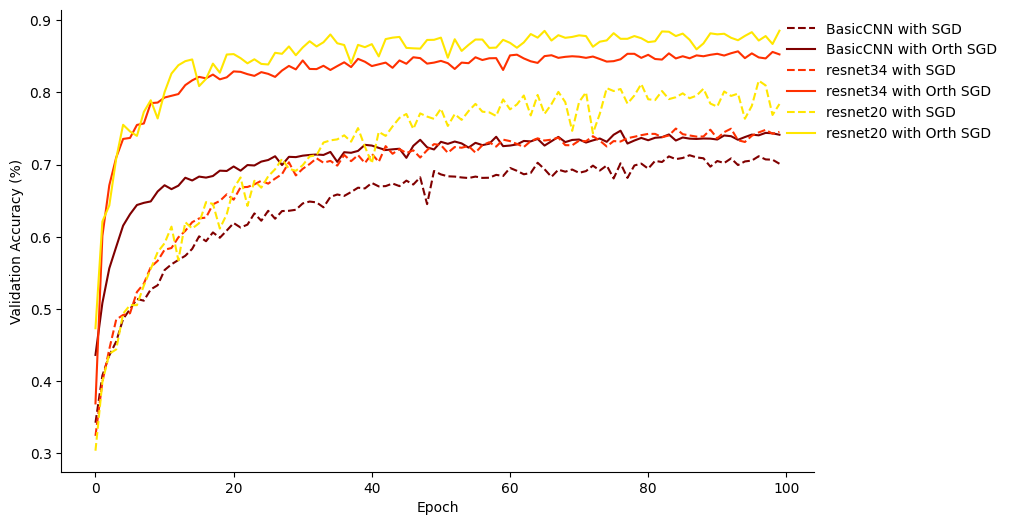}
	\caption{Orthogonalising just the convolutional filters vs both the convolutional layers and final dense layer on CIFAR-10; trained as in \cref{sec:cifar10}.}\label{fig:orth_conv_vs_dense_acc}
\end{figure}

\begin{table}[ht]
	\centering
	\caption{Test accuracy for \acrshort{osgdm} on CIFAR-10 when orthogonalising all layers vs orthogonalising just the convolutional layers. Trained as in \cref{sec:cifar10}, standard error across five runs.}\label{tab:dense_vs_conv_acc}
	\footnotesize
	\setlength\tabcolsep{2pt}
	\begin{tabular}{lccc}
		         & \acrshort{sgdm}          & \acrshort{osgdm}         & \shortstack{Conv \\ \acrshort{osgdm}}             \\
		\midrule
		BasicCNN & 73.60 {\tiny $\pm$ 0.19} & 76.67 {\tiny $\pm$ 0.10} & \textbf{76.80} {\tiny $\pm$ 0.18} \\
		resnet34 & 75.86 {\tiny $\pm$ 0.26} & 85.42 {\tiny $\pm$ 0.33} & \textbf{85.68} {\tiny $\pm$ 0.21} \\
		resnet20 & 79.14 {\tiny $\pm$ 0.62} & 87.12 {\tiny $\pm$ 0.12} & \textbf{87.70} {\tiny $\pm$ 0.40} \\
	\end{tabular}
\end{table}

\begin{table}[ht]
	\centering
	\caption{Test loss for \acrshort{osgdm} on CIFAR-10 when orthogonalising all layers vs orthogonalising just the convolutional layers. Trained as in \cref{sec:cifar10}, standard error across five runs.}\label{tab:dense_vs_conv_loss}
	\footnotesize
	\setlength\tabcolsep{2pt}
	\begin{tabular}{lcccccc}
		         & \acrshort{sgdm}             & \acrshort{osgdm}            & \shortstack{Conv \\ \acrshort{osgdm}}       \\
		\midrule
		BasicCNN & 0.7603 {\tiny $\pm$ 0.0061} & 0.6808 {\tiny $\pm$ 0.0038} & 0.6732 {\tiny $\pm$ 0.0041} \\
		resnet34 & 1.0468 {\tiny $\pm$ 0.0134} & 0.7087 {\tiny $\pm$ 0.0165} & 0.6268 {\tiny $\pm$ 0.0105} \\
		resnet20 & 0.6728 {\tiny $\pm$ 0.0301} & 0.6766 {\tiny $\pm$ 0.0155} & 0.4824 {\tiny $\pm$ 0.0225} \\
	\end{tabular}
\end{table}

As noted in \cref{sec:implementation} the extra wall time is dominated by the largest parameter, this is often the dense layer; \cref{tab:dense_vs_conv_acc,tab:dense_vs_conv_loss} show that for CIFAR-10 there is no impact on the error rate from not orthogonalising the final dense layer, and the training curves are the same shape --- \cref{fig:orth_conv_vs_dense_acc}.
While both the error rates and losses decrease when not orthogonalising the dense layer we hesitate to say that orthogonalising dense layers is detrimental since these networks only have a dense final classification layer which is qualitatively different from intermediary dense layers.

\subsection{Limitations due to mini-batch size}\label{sec:limitations}

\begin{figure}[!ht]
	\centering
	\includegraphics[width=0.9\linewidth]{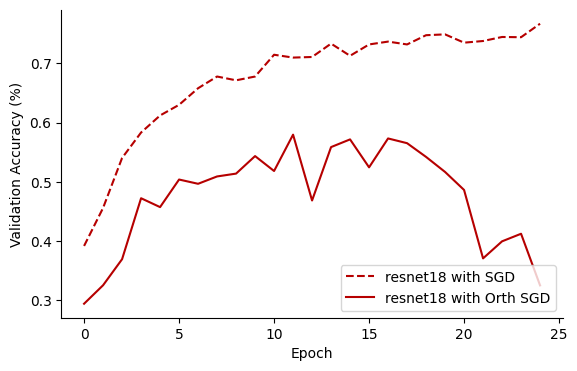}
	\caption{CIFAR-10 with mini-batch size=4 trained as in \cref{sec:cifar10}.}\label{fig:small_batch}
\end{figure}

\Gls{osgdm} does not perform as well as \gls{sgdm} when the mini-batch size is extremely small, \cref{fig:small_batch}, due to the increased levels of noise for the \gls{svd}.
This is the most likely reason that the resnet20 from \cref{sec:match_resnet} fails to match the original performance.

A mini-batch size of 16 is where \gls{osgdm} starts to outperform \gls{sgdm} on a resnet18 for CIFAR-10.
Few models need such small mini-batch sizes, but if they do then \gls{sgdm} would be a more suitable optimisation algorithm.
In addition to the learning collapse, the time taken by \gls{svd} is only dependent on the parameter size and not the mini-batch size, so increasing the number of mini-batches per epoch also increases the wall time to train.
The reason for the collapse in training with small mini-batch sizes will be subject to further research.

\section{Conclusion}
In this work we have laid out a new optimisation method, tested it on different models and data sets, showing close to state-of-the-art results out of the box and robustness to hyper-parameter choice and over-parametrised models.
\Gls{osgdm} also has practical application in problems such as object detection and semantic segmentation since they make use of a pre-trained image classification backbone.

\Gls{sgdm} with a vast amount of hyper-parameter tuning still reigns supreme, but \gls{osgdm} is an excellent method for quick verification of models or for prototyping --- when we want decent results fast, but do not need the absolute best performing model.
However, as more data set sizes are growing more models are being trained on fewer to less than one epoch of data~\citep{brown2020language} leading to an extremely limited ability to tune the hyper-parameters.

Lastly, we mentioned briefly in \cref{sec:intro} how attention heads fit our model but, since they are beyond the scope of this work, we will explore the potential gain in using \gls{osgdm} with them in future work, and expect a similarly exceptional gain will be obtained.

\bibliography{main}
\bibliographystyle{map/icml2022}

\newpage
\appendix
\onecolumn





\section{Cosine threshold}\label{sec:cosine_threshold}

We use the cosine metric
\begin{equation}
    \langle \vec{x}, \vec{y} \rangle_2 = \frac{\vec{x}\cdot \vec{y}}{\norm{\vec{x}}_2 \norm{\vec{y}}_2}
\end{equation}
in this work since it allows the comparison of the directions of high-dimensional vectors, and so obtain insight about the surface we are optimising on.
However, we note that the ``significance'' of the cosine between two random vectors depends on their size.

Assuming both vectors' components are random variables
\[
    x = \left[ x_1, x_2, \dots ,x_N \right],
\]
where $x_i\sim \mathcal{N}(0,\, \sigma^2)$, then the components of their dot product
\[
    \left\langle x,y \right\rangle = \left[ x_1\cdot y_1, x_2\cdot y_2, \dots x_N\cdot y_N \right],
\]
have variance $\sigma^4$.
Now, from the central limit theorem, the dot product has variance $\frac{\sigma^4}{N}$ where $N$ is the size of the vectors.
Finally, dividing by the magnitude of the vectors gives a variance of $\bar{\sigma} = \frac{\sigma^2}{N}$ for the cosine metric.

To gain some understanding of the significance of a cosine distance, we define a four-sigma threshold on the distribution of cosines, so, assuming $\sigma = 1, \mu = 0$, we get a threshold value of
\begin{align}
    \mu \pm 4\bar{\sigma} & = 0 \pm \frac{4\sigma}{\sqrt{N}} \nonumber \\
                          & = \pm \frac{4}{\sqrt{N}}
    \label{eq:random_vec_threshold}
\end{align}

This is important because a distance of $0.1$ might seem small, but for $10,000$-dimensional vectors, it easily clears our significance threshold.

\section{Model Summaries}
\subsection{BasicCNN}\label{sec:BasicCNN}
\begin{lstlisting}[language={}]
----------------------------------------------------------------
        Layer (type)               Output Shape         Param #
================================================================
            Conv2d-1           [-1, 32, 16, 16]             896
       BatchNorm2d-2           [-1, 32, 16, 16]              64
            Conv2d-3             [-1, 32, 8, 8]           9,248
       BatchNorm2d-4             [-1, 32, 8, 8]              64
            Conv2d-5             [-1, 32, 4, 4]           9,248
       BatchNorm2d-6             [-1, 32, 4, 4]              64
            Linear-7                   [-1, 10]           5,130
          BasicCNN-8                   [-1, 10]               0
================================================================
Total params: 24,714
Trainable params: 24,714
Non-trainable params: 0
----------------------------------------------------------------
Input size (MB): 0.01
Forward/backward pass size (MB): 0.16
Params size (MB): 0.09
Estimated Total Size (MB): 0.27
----------------------------------------------------------------
\end{lstlisting}

\clearpage
\section{Full results plot}\label{sec:full_orth}
\begin{figure}[h!t]
    \centering
    \begin{subfigure}{\textwidth}
        \centering
        \includegraphics[width=\linewidth]{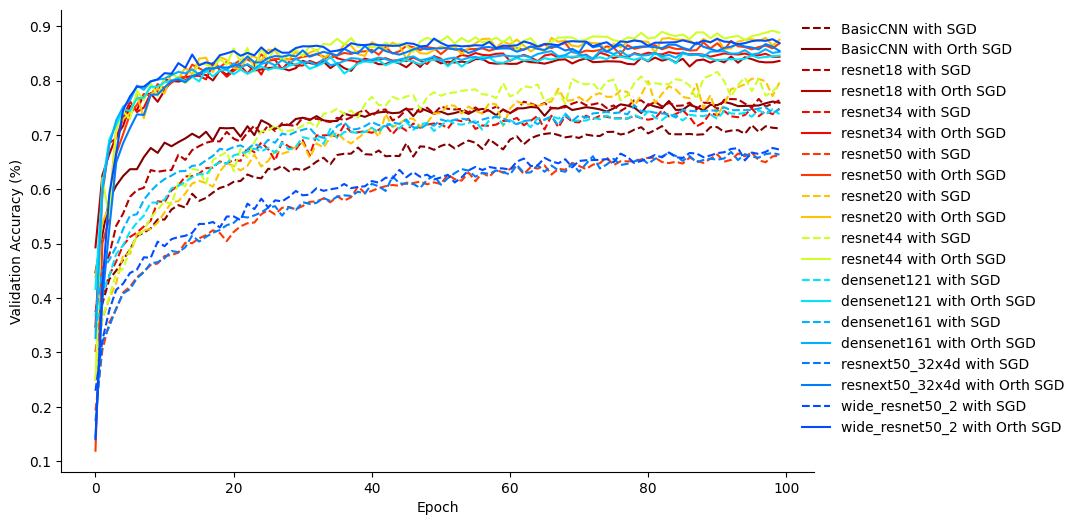}
        \caption{Validation Accuracy}
    \end{subfigure}
    \begin{subfigure}{\textwidth}
        \centering
        \includegraphics[width=\linewidth]{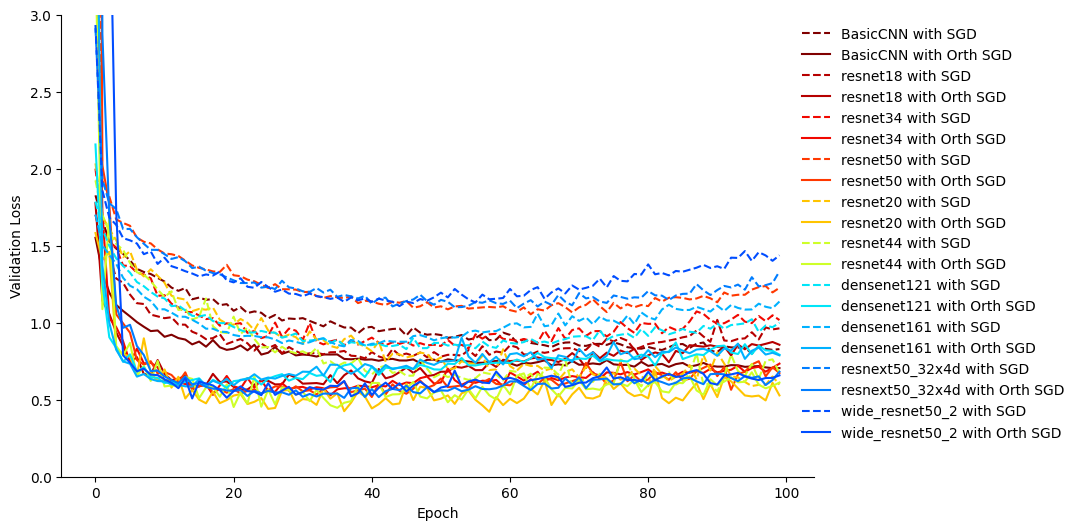}
        \caption{Validation Loss}
    \end{subfigure}
    \caption{\gls{sgdm} vs Orthogonal \gls{sgdm}}\label{fig:sgd_vs_orth_sgd_all}
\end{figure}

\begin{table}[ht]
    \centering
    \caption{Test loss across a suite of hyper-parameter sets on CIFAR-10 on a resnet20, standard error across five runs. For \gls{adam} $\beta_2 = 0.99$.}\label{tab:orth_cifar_results_adam_loss}
    \begin{tabular}{llcccc}
                  &          & \multicolumn{2}{c}{\gls{sgdm}}        & \multicolumn{2}{c}{\gls{adam}}                                                                                       \\
        \cmidrule(lr){3-4}\cmidrule(lr){5-6}
        LR        & Momentum & Original                              & Orthogonal                            & Original                             & Orthogonal                            \\ \midrule

        $10^{-1}$ & 0.95     & 0.5487 {\tiny $\pm$ 0.0913 }          & \textbf{0.4916} {\tiny $\pm$ 0.0420 } & 2.4148 {\tiny $\pm$ 0.7090 }         & 0.7938 {\tiny $\pm$ 0.0663 }          \\
        $10^{-2}$ & 0.95     & 0.5762 {\tiny $\pm$ 0.0431 }          & \textbf{0.4772} {\tiny $\pm$ 0.0277 } & 0.8395 {\tiny $\pm$ 0.1079 }         & 0.5415 {\tiny $\pm$ 0.0153 }          \\
        $10^{-3}$ & 0.95     & 0.9323 {\tiny $\pm$ 0.0149 }          & \textbf{0.4988} {\tiny $\pm$ 0.0114 } & 0.6013 {\tiny $\pm$ 0.0305 }         & 0.5998 {\tiny $\pm$ 0.0131 }          \\
        \midrule
        $10^{-1}$ & 0.9      & 0.6089{\tiny $\pm$ 0.0535}            & \textbf{0.5370}{\tiny $\pm$ 0.0230}            & 9.0193{\tiny $\pm$ 5.9607}           & 0.9803{\tiny $\pm$ 0.1607}            \\
        $10^{-2}$ & 0.9      & 0.6217{\tiny $\pm$ 0.0194}            & \textbf{0.4841}{\tiny $\pm$ 0.0087}            & 0.8898{\tiny $\pm$ 0.0447}           & 0.5903{\tiny $\pm$ 0.0201}            \\
        $10^{-3}$ & 0.9      & 1.0977{\tiny $\pm$ 0.0036}            & \textbf{0.5042}{\tiny $\pm$ 0.0055}            & 0.6045{\tiny $\pm$ 0.0312}           & 0.5829{\tiny $\pm$ 0.0102}            \\
        \midrule
        $10^{-1}$ & 0.8      & \textbf{0.5395} {\tiny $\pm$ 0.0091 } & 0.5414 {\tiny $\pm$ 0.0459 }          & 4.5143 {\tiny $\pm$ 1.6628 }         & 1.0984 {\tiny $\pm$ 0.2710 }          \\
        $10^{-2}$ & 0.8      & 0.6669 {\tiny $\pm$ 0.0290 }          & \textbf{0.4940} {\tiny $\pm$ 0.0163 } & 0.8741 {\tiny $\pm$ 0.0627 }         & 0.5114 {\tiny $\pm$ 0.0204 }          \\
        $10^{-3}$ & 0.8      & 1.2805 {\tiny $\pm$ 0.0106 }          & \textbf{0.5238} {\tiny $\pm$ 0.0034 } & 0.6950 {\tiny $\pm$ 0.0656 }         & 0.5671 {\tiny $\pm$ 0.0088 }          \\
        \midrule
        $10^{-1}$ & 0.5      & 0.6796 {\tiny $\pm$ 0.0158 }          & \textbf{0.5003} {\tiny $\pm$ 0.0148 } & 15683.8291 {\tiny $\pm$ 15681.5332 } & 0.9481 {\tiny $\pm$ 0.1072 }          \\
        $10^{-2}$ & 0.5      & 0.8927 {\tiny $\pm$ 0.0304 }          & \textbf{0.4950} {\tiny $\pm$ 0.0121 } & 0.9594 {\tiny $\pm$ 0.0846 }         & 0.6610 {\tiny $\pm$ 0.0285 }          \\
        $10^{-3}$ & 0.5      & 1.5309 {\tiny $\pm$ 0.0189 }          & 0.6284 {\tiny $\pm$ 0.0186 }          & 0.6482 {\tiny $\pm$ 0.0511 }         & \textbf{0.6228} {\tiny $\pm$ 0.0213 } \\
    \end{tabular}
\end{table}

\begin{figure}[h!t]
    \centering
    \begin{subfigure}{\textwidth}
        \centering
        \includegraphics[width=\linewidth]{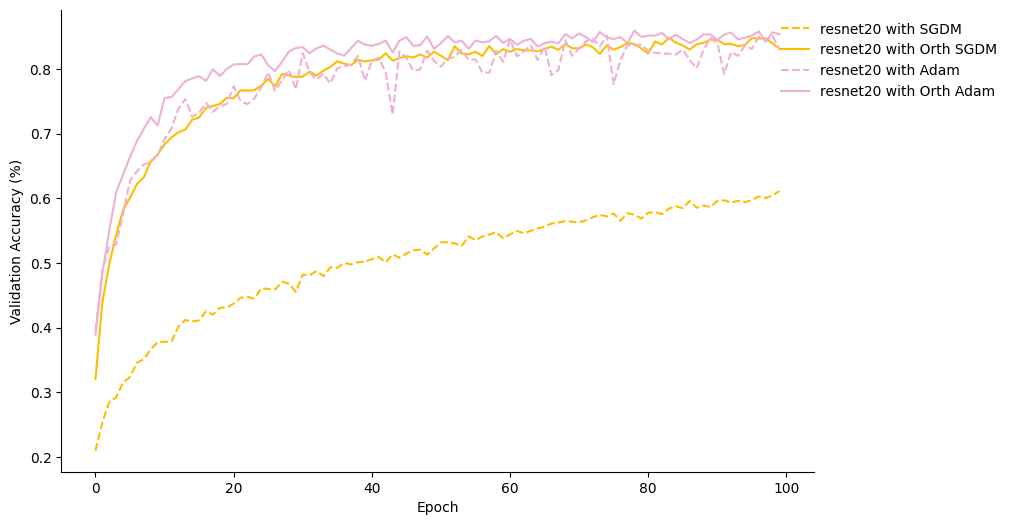}
        \caption{Validation Accuracy}
    \end{subfigure}
    \begin{subfigure}{\textwidth}
        \centering
        \includegraphics[width=\linewidth]{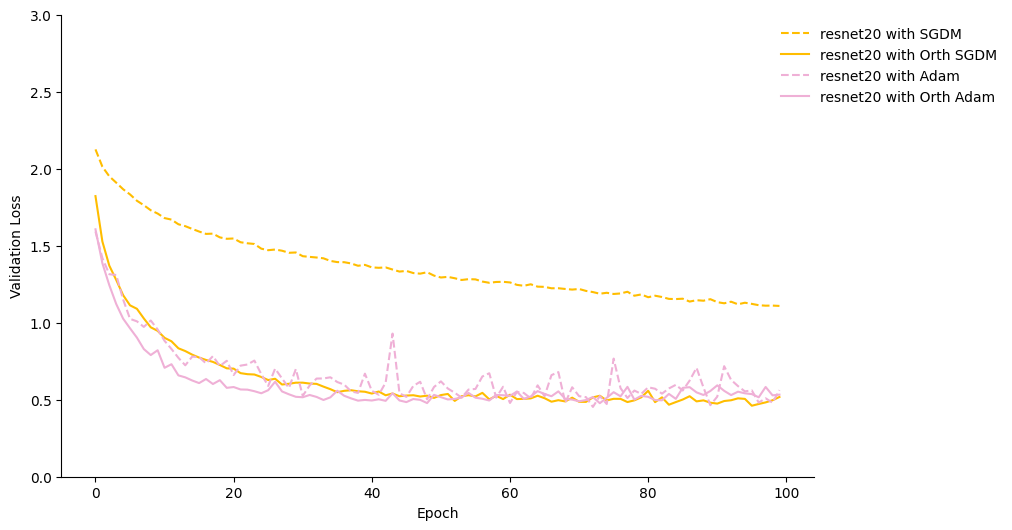}
        \caption{Validation Loss}
    \end{subfigure}
    \caption{A compassion of \gls{adam} and \acrshort{sgdm}, learning rate = $1\times 10^{-3}$, $\beta_1 = 0.9, \: \beta_2 = 0.99$.}\label{fig:adam1}
\end{figure}

\begin{figure}[h!t]
    \centering
    \begin{subfigure}{\textwidth}
        \centering
        \includegraphics[width=\linewidth]{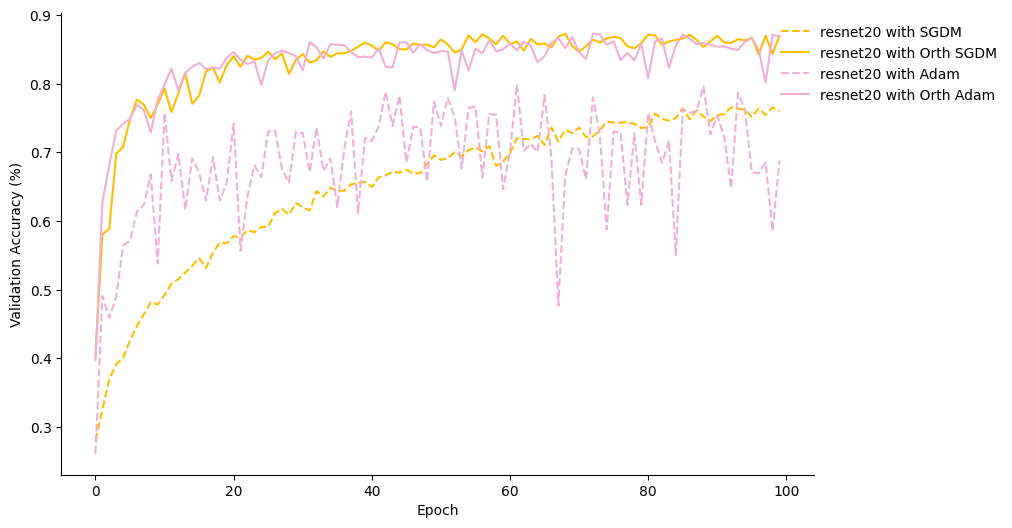}
        \caption{Validation Accuracy}
    \end{subfigure}
    \begin{subfigure}{\textwidth}
        \centering
        \includegraphics[width=\linewidth]{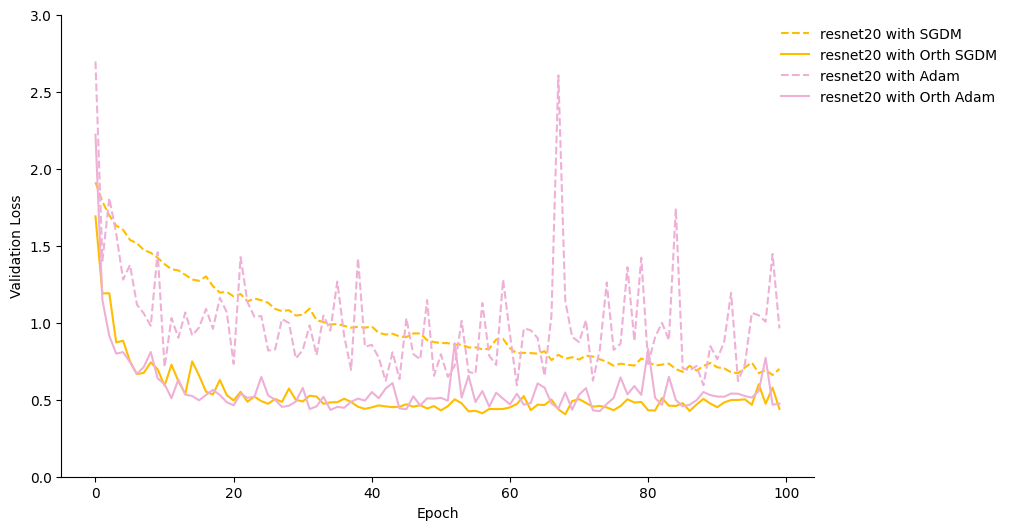}
        \caption{Validation Loss}
    \end{subfigure}
    \caption{A compassion of \gls{adam} and \acrshort{sgdm}, learning rate = $1\times 10^{-2}$, $\beta_1 = 0.8, \: \beta_2 = 0.99$.}\label{fig:adam2}
\end{figure}

\begin{table}[ht]
    \centering
    \caption{Test loss for several models trained with \gls{sgdm}, \acrlong{nsgdm}, \acrlong{cnsgdm}, and \gls{osgdm}.}\label{tab:normalised_sgdm_loss}
    \begin{tabular}{lcccc}
                    & \acrshort{sgdm}                     & \acrshort{nsgdm}           & \acrshort{cnsgdm}          & \acrshort{osgdm}                    \\ \midrule
        BasicCNN    & 0.7559{\tiny $\pm$ 0.0065}          & 0.7637{\tiny $\pm$ 0.0098} & 0.7443{\tiny $\pm$ 0.0094} & \textbf{0.6824}{\tiny $\pm$ 0.0081} \\
        resnet18    & 0.9252{\tiny $\pm$ 0.0098}          & 0.9726{\tiny $\pm$ 0.0214} & nan{\tiny $\pm$ nan}       & \textbf{0.7938}{\tiny $\pm$ 0.0083} \\
        resnet50    & 1.0950{\tiny $\pm$ 0.0181}          & 0.9454{\tiny $\pm$ 0.0126} & nan{\tiny $\pm$ nan}       & \textbf{0.6785}{\tiny $\pm$ 0.0076} \\
        resnet44    & \textbf{0.7093}{\tiny $\pm$ 0.0678} & 0.7641{\tiny $\pm$ 0.0441} & 0.7902{\tiny $\pm$ 0.0477} & 0.7694{\tiny $\pm$ 0.0426}          \\
        densenet121 & 0.9357{\tiny $\pm$ 0.0071}          & 1.0096{\tiny $\pm$ 0.0167} & nan{\tiny $\pm$ nan}       & \textbf{0.8142}{\tiny $\pm$ 0.0084} \\
    \end{tabular}
\end{table}

\end{document}